\pdfoutput=1

\documentclass[11pt]{article}

\usepackage{emnlp2021}

\usepackage{times}
\usepackage{latexsym}
\usepackage{multirow}
\usepackage{hyperref}
\usepackage{enumitem}

\usepackage[T1]{fontenc}
\usepackage{float}
\usepackage{subcaption}
\usepackage{graphicx}

\usepackage[utf8]{inputenc}

\usepackage{microtype}

\title{What BERT Based Language Models Learn in Spoken Transcripts: An Empirical Study}

\author{Ayush Kumar, Mukuntha Narayanan Sundararaman \and Jithendra Vepa \\
        \texttt{\{ayush, mukunthas, jithendra\}@observe.ai}\\
        Observe.AI \\ Bangalore, India}

\begin{document}
\maketitle
\begin{abstract}
Language Models (LMs) have been ubiquitously leveraged in various tasks including spoken language understanding (SLU). 
Spoken language requires careful understanding of speaker interactions, dialog states and speech induced multimodal behaviors to generate a meaningful representation of the conversation. In this work, we propose to dissect SLU into three representative properties: \textit{conversational} (disfluency, pause, overtalk), \textit{channel} (speaker-type, turn-tasks) and \textit{ASR} (insertion, deletion, substitution). 
We probe BERT based language models (BERT, RoBERTa) trained on spoken transcripts to investigate its ability to understand multifarious properties in absence of any speech cues.
Empirical results indicate that LM is surprisingly good at capturing conversational properties such as pause prediction and overtalk detection from lexical tokens. 
On the downsides, the LM scores low on turn-tasks and ASR errors predictions.
Additionally, pre-training the LM on spoken transcripts restrain its linguistic understanding. Finally, we establish the efficacy and transferability of the mentioned properties on two benchmark datasets: Switchboard Dialog Act and Disfluency datasets.
\end{abstract}

\section{Introduction}
\label{sec:intro}

Language Models (LM) such as BERT, GPT etc., have led to
significant advancements in the field of Natural Language Processing (NLP) by learning representations pre-trained on a vast
amount of text data. These LMs have shown to perform well on a
multitude of downstream tasks such as question answering, intent, entity and sentiment detection, natural language inference
etc. \cite{DBLP:conf/naacl/DevlinCLT19}.
Along with NLP, the proliferation of voice-enabled technologies has resulted in the significance of spoken language understanding (SLU) tasks. 
The general modus-operandi of SLU systems is to convert voice into text using an ASR engine and use natural language understanding (NLU) on the transcribed text 
by modelling \textit{conversational} and \textit{channel} properties while being robust to \textit{ASR} errors. 
Since spoken conversation boasts of amalgamation of spontaneous speaker interactions, it has become imperative for model architectures to capture multimodal features from text and speech modalities \cite{DBLP:conf/interspeech/GeorgiouPP19}.
The aim of these multimodal systems is to capture acoustic information such as pitch, intonation, rate of speech, etc. along with properties such as pause, disfluency, overtalk and turn shifts to synthesize a complete understanding of what is being spoken and how it is being spoken. 
While some of these properties rely on speech signals and can only be captured by acoustic systems, properties such as pauses, disfluency, overtalk, turn shifts, speech recognition errors etc. get passed onto the lexical domain in the form of spoken transcripts and meta-information.
Few works suggest that the auxiliary tasks such as speaker intent detection \cite{DBLP:conf/icassp/AldenehDP18}, filler prediction \cite{DBLP:conf/interspeech/HaraITK18}, pause and speaking-rate \cite{DBLP:conf/interspeech/WeinerES17} help to train a robust model for downstream tasks. 
Thus, there is a need for a \textit{vigilant} language model capable of encoding the multi-dimensional properties of a SLU dataset to reduce dependence on auxiliary information and/or multimodal inputs. However, in order to come up with such LM for SLU domain, a pre-requisite is to investigate the extent to which an LM already encodes these properties.

In this work, we propose to categorize the spoken language understanding into distinctive properties:
1) \textit{Conversational}: encompass spontaneous speech properties such as pauses, disfluency, overtalk; 2) \textit{Channel}: properties that capture speaker turns such as speaker-type, turn-taking; 3) \textit{ASR}: properties such as error type (insertion, deletion, substitution). While core focus in this work is to not propose a state-of-the-art language model, we believe that unveiling the behavior of LM on these properties will not only help us to understand the encoded information in a pre-trained language model, but will also influence architectural designs to build upon the findings. Thus, \textit{we limit our experiments to utilize only lexical tokens present in the spoken transcripts}. Specifically, we aim to answer the following research questions (RQ) in our work:
\vspace{-5pt}
\begin{itemize}
    \item \textbf{RQ1}: Can language model encode the distinctive properties of SLU datasets in absence of any speech cues?
    \begin{itemize}
        \item \textit{RQ1a}: Does LM representations capture conversational properties such as disfluency, overtalk, pause?
        \item \textit{RQ1b}: How well can LM encapsulate channel understanding: speaker identity of a channel and turn-tasks?
        \item \textit{RQ1c}: Can LM identify insertion, substitution and deletion errors in the spoken transcripts and encode word error rate?
    \end{itemize}
    \vspace{-5pt}
    \item \textbf{RQ2}: How much of linguistic understanding is retained by a LM trained on noisy spoken transcripts?
    \vspace{-5pt}
    \item \textbf{RQ3}: Can an LM infused with such properties perform better on an unseen spoken language dataset compared to a vanilla LM?
\end{itemize}





To answer RQ1, we frame a set of probing tasks (Table \ref{examples}, Section \ref{probing}). Probing tasks are a introspection mechanism to unearth the properties encoded in the neural model (LM) \cite{DBLP:conf/iclr/AlainB17, DBLP:conf/iclr/TenneyXCWPMKDBD19}. Depending on the task, a probe model can be a classification or a regression setup that is trained on LM-extracted features. The weights of LM are frozen while training the probe model. If the result on a probing task is good enough, we say that the property under the scanner is encoded in the neural model. For RQ2, we evaluate LM on a couple of linguistic properties: dependency relations and entity-value mapping. Finally, we infuse the conversational, channel and ASR properties via a multi-task (MTL) framework trained on probe tasks and evaluate the trained model on unseen spoken language tasks (RQ3).
The empirical results and analysis reveal the following insights:
\begin{enumerate}
    \item LM encodes the conversational properties such as pause and overtalk surprisingly well (\textit{RQ1a}). Channel properties are captured well enough for only speaker role prediction while LM finds itself struggling to encode information relevant for turn-tasks (\textit{RQ1b}). ASR properties are poorly captured in the LM. While the model can detect if a token is erroneous, it is inaccurate in predicting the error type. Interestingly, ASR transcript pre-trained LM detects substitution much better than vanilla LM in comparison to insertion and deletion errors (\textit{RQ1c}).
    \vspace{-8pt}
    \item Once trained on spoken transcripts, the LM vastly loses its understanding of dependency relations. Model also scores poorly on entity-value mapping (\textit{RQ2}).
    \vspace{-8pt}
    \item An LM fine-tuned on probing tasks with a multi-task learning (MTL) framework results in an improved performance on Switchboard Dialog Act and Disfluency corpus. This shows that language model representations which encode such properties perform better on downstream tasks (\textit{RQ3}).
\end{enumerate}

\section{Probing Tasks (RQ1)}
\label{probing}

In this section, we define probing tasks with respective examples. We also describe the dataset creation methodology in detail in Section \ref{creation}.


\begin{table*}[t]
\small
\centering
\renewcommand{\arraystretch}{1.05}
\begin{tabular}{c|l}
\textbf{Property Type}                                         & \textbf{Probe Tasks}                                                                                                                                                                                                                \\ \hline
\multicolumn{1}{l|}{\multirow{17}{*}{\textbf{Conversational}}} &                                                                                                                                                                                                                                     \\
\multicolumn{1}{l|}{}                                          & \textbf{Disfluency Detection: Is the given spoken utterance disfluent?}                                                                                                                                                             \\
\multicolumn{1}{l|}{}                                          & \textit{no so no i i don't can i have instead get a refund}                                                                                                                                                                         \\
\multicolumn{1}{l|}{}                                          & \textit{well i don't have a i mean i probably do have a provider there}                                                                                                                                                             \\
\multicolumn{1}{l|}{}                                          &                                                                                                                                                                                                                                     \\
\multicolumn{1}{l|}{}                                          & \textbf{Pause Identification: Does the speaker take long pauses while speaking?}                                                                                                                                                    \\
\multicolumn{1}{l|}{}                                          & \textit{let’s see \textless{}silence\textgreater{}i see so we haven’t tried to go on the website}                                                                                                                                   \\
\multicolumn{1}{l|}{}                                          & \textit{okay \textless{}silence\textgreater{}so what i see is that your request is already with us}                                                                                                                                 \\
\multicolumn{1}{l|}{}                                          &                                                                                                                                                                                                                                     \\
\multicolumn{1}{l|}{}                                          & \textbf{Overtalk Detection: Are two speakers talking over each other?}                                                                                                                                                              \\
\multicolumn{1}{l|}{}                                          & \textit{i’m {[}yes{]} not {[}i know{]} referring to {[}how{]} the {[}it works{]} transaction}                                                                                                                                       \\
\multicolumn{1}{l|}{}                                          & \textit{i {[}you{]} thought {[}want{]} that {[}me to{]} you {[}do{]} gotta {[}that{]} do {[}for you{]}}                                                                                                                             \\
\multicolumn{1}{l|}{}                                          &                                                                                                                                                                                                                                     \\
\multicolumn{1}{l|}{}                                          & \textbf{Question Classification: Did the speaker ask any question?}                                                                                                                                                                 \\
\multicolumn{1}{l|}{}                                          & \textit{{[}entity{]}: and your date of birth please}                                                                                                                                                                                \\
\multicolumn{1}{l|}{}                                          & \textit{{[}choice{]}: so you want to go with monthly plan or yearly}                                                                                                                                                                \\
\multicolumn{1}{l|}{}                                          &                                                                                                                                                                                                                                     \\ \hline
\multirow{13}{*}{\textbf{Channel}}                             &                                                                                                                                                                                                                                     \\
                                                               & \textbf{Speaker Role: Who is the speaker for a given utterance?}                                                                                                                                                                    \\
                                                               & \textit{agent: you can cancel anytime you are not obligated to write to us}                                                                                                                                                         \\
                                                               & \textit{customer: thank you i tried yesterday i was told to call back today}                                                                                                                                                        \\
                                                               &                                                                                                                                                                                                                                     \\
                                                               & \textbf{Response Length: Is the expected response to current utterance is short or long?}                                                                                                                                           \\
                                                               & \textit{short: \textless{}can i get you full name please\textgreater{}\textless{}it’s johnson rest\textgreater{}}                                                                                                                   \\
                                                               & \textit{long: \textless{}may i know what didn’t work\textgreater{}\textless{}when i came to webpage ...\textgreater{}}                                                                                                              \\
                                                               &                                                                                                                                                                                                                                     \\
                                                               & \textbf{Turn Taking: Has speaker completed its turn?}                                                                                                                                                                               \\
                                                               & \textit{turn-break: \textless{}appreciate your reply i’ll get that done for you\textgreater{}\textless{}silence\textgreater{}}                                                                                                      \\
                                                               & \textit{\begin{tabular}[c]{@{}l@{}}turn-continue: \textless{}i get that let me verify that for you \textless{}silence\textgreater{}okay i’ve\\ pushed data from my end you should hear back in 48 hours\textgreater{}\end{tabular}} \\
                                                               &                                                                                                                                                                                                                                     \\ \hline
\multirow{10}{*}{\textbf{ASR Errors}}                          &                                                                                                                                                                                                                                     \\

                                                               & \textbf{Error Type: What category of ASR error does a particular token belong to?}                                                                                                                                                  \\
                                                               & \textit{reference: \textless{}customer resolution is our primary motive\textgreater{}}                                                                                                                                              \\
                                                               & \textit{hypothesis: \textless{}customer resolution is hour primary motive\textgreater{}}                                                                                                                                            \\
                                                               & \textit{hour: substitution error} \\                         
                                                           &     \\
                                                               
                                                               & \textbf{WER Score: What is the word-error-recognition score for the transcript?}                                                                                                                                                    \\
                                                               & \textit{reference: \textless{}customer resolution is our primary motive\textgreater{}}                                                                                                                                              \\
                                                               & \textit{hypothesis: \textless{}customer resolution is hour primary motive\textgreater{}}                                                                                                                                            \\
                                                               & \textit{WER: 16.67}                                                                                                                                                                                                                 \\
                                                               &                                                                                                                                                                                                 
\end{tabular}
\caption{Representative examples for each probing task. Special tokens or markers are pruned before running the probe setup.}
\vspace{-1.5em}
\label{examples}
\end{table*}

\subsection{Conversational Properties (RQ1a)}

\subsubsection{Disfluency Detection}
An important characteristic of spontaneous speech that makes it distinguishable from a written text or formal speech is the presence of disfluencies. Disfluency can refer to filler words (\textit{um, uh}), repetitions (\textit{this this is just not working}), false starts (\textit{you were okay that's perfectly fine}), discourse markers (\textit{i mean}) etc. that are an integral part of natural conversations. Despite disfluencies considered as noise to downstream NLU tasks, they indicate distinguishing behaviors such as low confidence of the speaker \cite{DBLP:conf/icassp/DinkarVPC20} and non-native speaker of the language. Hence, identification of disfluent text in spoken transcripts is of prime importance to capture auxiliary speaker behaviors.

In this work, we create a disfluency dataset that has the instances categorized into fluent vs disfluent classes. A probe classifier is set up on this dataset for disfluency detection.

\subsubsection{Pause Identification}


Pauses and silences are natural elements of human interaction. Any spontaneous conversation would contain participants taking pauses and silences indicating situations such as: holds for retrieving  information, transition to a new topic of discussion, silence to gather thoughts, etc. Identification of pauses can make the language model understand the current dialog state \textit{viz.,} information exchange, call-hold, turn-shifts, topic-change etc.

In this work, we treat silences longer than 5 sec. as a pause and pose pause identification as a binary classification problem with an intent to understand how well an LM encodes the information about the presence or absence pause without taking any speech cues. 

\subsubsection{Overtalk Detection}
Overtalk is a phenomenon that occurs when multiple parties are speaking at the same time. While overtalk is inevitable in a natural conversation, too much overtalk in conversations indicate poor listening skills displayed by the involved parties. Detecting overtalk is straightforward in dual-channel (diarized) calls, but in mono-channel (non-diarized) calls the speaker segmentation is not available. Although diarization is a preferred way to identify speakers, such models are non-trivial to train and would contain margin of errors.

We identify overtalks in dual-channel calls and convert the corresponding spoken transcript in a sequence of time-ordered tokens (mono-channel transcript) for the overtalk detection task (Table \ref{examples}). Overtalk detection is a binary classification task where one class contains text that contains overtalk between speakers while the other is a turn uttered by single speaker.

\subsubsection{Question Identification}
An interaction between an agent and a customer hinge around multiple information exchanges often anchored via a question-answer pair. The intent of the question, eg: entity, descriptive, boolean etc. additionally helps to track the anatomy of the call which captures activities such as agent verifying customers (entity questions), agent probing customers to gather complete understanding of the issue (descriptive questions) etc. However, in spoken language, the formal structure of the question is not always followed and often questions are asked in non-surface forms (\textit{and your account number}) instead of \textit{can you provide your account number}). Thus, question identification is a challenging task in SLU domain. We frame question identification as a multi-class setup with four classes: entity, descriptive, boolean and choice based.

\subsection{Channel Properties (RQ1b)}

\subsubsection{Speaker Role}
Identity of the current speaker and speaker roles (sender, addressee,  observer, etc.) facilitates a better understanding of the dialog state in a conversation \cite{kim2020speaker}. Speaker roles and identities have also been leveraged in context extraction and response generation in dialog systems \cite{DBLP:conf/aaai/ZhangLPR18}. Thus, we formulate a probe task to identify the role of the current speaker in a call. In a call center conversation, there are three speaker roles that co-exist: agent, customer and IVR. We formulate a multiclass classification task to probe the LM on identifying speaker roles.

\subsubsection{Turn Tasks}
Understanding contexts (turns) in a conversation is a key step to frame a better understanding of the conversation. To probe on how well the LM encode the contexts, we posit two turn-tasks:
\vspace{-5pt}
\begin{itemize}
    \item \textit{Response-Length}: We set up a binary classification task where the probe classifier predicts the duration of the next-turn (\textit{short} or \textit{long}) given the current and previous turn. This task is an interesting intersection of spoken and written properties of SLU where a turn that is as short as (\textit{let me see. oh seems you're correct}) in the lexical domain can be as long as a minute in the time domain because of the silence that speaker took while looking-up the information. Hence, we aim to probe LM on not just lexical understanding of tokens but also identify if LM captures such subtleties of spoken transcripts. We consider any turn that lasts up to 30 seconds as a short-response and a turn lasting longer than that as a long-response.
    \vspace{-5pt}
    \item \textit{Turn-Taking}: In a real-time system, the spoken transcript comes in as a batch of tokens. In such a system, it becomes important to understand if the speaker has completed its turn and is ready to listen. Inspired by \cite{DBLP:conf/interspeech/RoddySH18}, we formulate the situation as a binary text classification where the task is to predict if the speaker would continue speaking at the end of the utterance or would it handover the turn to the other party. We use the pause markers in the dataset to identify the discourse segments which is used to generate the dataset for turn-taking.
\end{itemize}

\subsection{ASR Properties  (RQ1c)}

\subsubsection{Error Type Prediction}
We formulate this task as a token classification task i.e, a single token of an entire utterance is considered at a time for probing. We define two setups for error type prediction: a) \textit{Binary Classification}: Given a token in spoken transcript, classify if the token is correct or erroneous; b) \textit{Multiclass Classification}: Given an erroneous token in spoken transcript, classify the error type of the token. In case of deletion, the token next to deleted token is probed for the error type. We believe this is an ideal setup as in real scenario, one can only distinguish a deleted token only when next token is predicted in the ASR transcript.

\subsubsection{Word Error Rate (WER) Prediction}
In downstream tasks such as call summarization, the dialog turns that are highly noisy (contains ASR errors) are undesirable and should not be focused much while generating summaries. To achieve this property, it becomes inevitable for the model to identify such dialog turns with high WERs. Since any such recent summarization system uses language modelling as underlying component, we probe the LM on its ability to predict the WER of a given dialog turn.

\section{Linguistic Understanding of LM (RQ2)} 

Deep language models such as BERT have been shown to encode a range of syntactic and semantic information with more complex structures represented hierarchically in the higher layers of the model \cite{DBLP:conf/acl/BaroniBLKC18, DBLP:conf/acl/JawaharSS19}.
\citet{DBLP:journals/corr/abs-2012-15150} emphasize on capturing syntactic structure via syntax aware local attention while \citet{DBLP:journals/corr/abs-1911-06156} propose injecting structural and syntactic information such as parts-of-speech tags which results in a higher performing machine translation model. 
Since spoken language transcripts contain spontaneous texts devoid of formal structure and incorrect grammars added with speech to text conversion errors, it becomes imperative to quantify the linguistic structures encoded in the language model. We probe LM for such properties via these setups:

\begin{enumerate}[noitemsep,topsep=0pt]
    \item \textbf{\textit{Dependency Parsing}}: Following the work in \citet{DBLP:clark}, we  investigate individual attention heads in both directions to probe what aspects of language they have retained after being pre-trained with spoken language transcripts on the benchmark task of dependency parsing. We use Penn Treebank dataset \cite{DBLP:journals/coling/MarcusSM94} tagged with Stanford's dependencies to report the performance on dependency parsing.
    
    \item \textbf{\textit{Entity-Value Mapping}}: In an agent-customer conversation, agent usually asks some PII entities to verify the identity of the customer before proceeding further into the call (eg: \textit{can you provide me your date of birth and last name}). The customer replies with a reference to both entities asked by the agent (eg: \textit{sure that would be davis and ninth march ninety four}
    For an entity extraction system to work in such spoken language transcripts, the model should be able to map \textit{davis} to \textit{last name} and \textit{ninth march ninety four} to \textit{date of birth}. In this work, we pose this problem as a entity-value mapping. Particularly, we compute entity selection accuracy measured by what percent of the time does the head word of a entity value most attend to the head of its entity. For example, if 3 tokens out of \textit{ninth march ninety four} attends maximum to any of the token in \textit{date of birth}, we would say that entity selection accuracy is 75\%.
\end{enumerate}

\section{Properties Infused LM: A Multi-Task Learning setup (RQ3)}

We hypothesize that an LM that better encodes the conversational, channel and ASR properties in its representation performs better on downstream SLU tasks. In order to verify the hypothesis, we follow a two step pipeline:
\begin{enumerate}[noitemsep,topsep=0pt]
    \item In the first step, we infuse LM with these properties by fine-tuning both LMs on all probe tasks simultaneously using a multi-task learning (MTL) framework.
    \item Next, we freeze the weights of the LMs and evaluate the performance of the probe classifier on a couple of external datasets: Switchboard Dialog Act and Switchboard disfluency for dialog act classification and disfluency detection respectively. It should be noted that none of these datasets are utilized in the pre-training step of the two LMs.
\end{enumerate}
Since MTL-trained LMs get explicit supervision to capture properties needed by probing tasks, their performance on downstream tasks would also indicate the relevance of the proposed probing tasks.



\section{Experimental Setup}
\label{exp}

\subsection{Language Model}
We use BERT \cite{DBLP:conf/naacl/DevlinCLT19} as our base language model. Since, RoBERTa \cite{DBLP:journals/corr/abs-1907-11692}, which is a robustly-optimized version of BERT is also trained on masked language modelling task and has been reported to outperform BERT on several NLP tasks, we extend our probe to RoBERTa. We observe a better result, in general, from RoBERTa-base model. Hence, we run further experiments on variations of RoBERTa-base model. 
Specifically, we compare following setups:
\begin{itemize}[leftmargin=0.25cm,labelwidth=\itemindent,labelsep=0.2cm,align=left,noitemsep,topsep=0pt]
    \item \textbf{\textit{BERT-base}}: We use pre-trained \textit{BERT-base-uncased} model to evaluate out-of-box performance on the probe tasks.
    \item \textbf{\textit{RoBERTa-base}}: Since, RoBERTa is a BERT-like model robustly trained with larger dataset, we also utilize pre-trained \textit{RoBERTa-base} model to report the results.
    \item \textbf{\textit{Chat-RoBERTa}}: We pre-train RoBERTa on utterances derived from two dialog corpus: PersonaChat \cite{DBLP:conf/acl/KielaWZDUS18} and MultiOz dataset \cite{DBLP:conf/emnlp/BudzianowskiWTC18}. The pre-training is done with MLM task to influence the LM with dialog properties, with an aim to establish a competitive baseline trained on a large spontaneous dialog dataset.
    \item \textbf{\textit{ASRoBERTa}}: It is imperative to pre-train LM on large domain data for a better performance. ASRoBERTa is pre-trained on in-domain ASR corpus to evaluate the impact of domain pre-training on the performance on probing tasks ($\Delta_{domain}$).
    \item \textbf{\textit{Oracle}}: In an ideal setup, it is suggested to fine-tune the LM on the downstream task for best results. We report results on task fine-tuned model (Oracle) to understand the limits of the LM in an ideal setup 
    compared to in-domain training.
\end{itemize}

\begin{table*}[]
\footnotesize
\renewcommand{\arraystretch}{1.25}
\begin{tabular}{l|l|cc|ccc|cc}
\multicolumn{2}{c|}{\textbf{Probing Task}}                       & \textbf{\begin{tabular}[c]{@{}c@{}}BERT\\ -base\end{tabular}} & \textbf{\begin{tabular}[c]{@{}c@{}}RoBERTa\\ -base\end{tabular}} & \textbf{\begin{tabular}[c]{@{}c@{}}Chat\\ -RoBERTa\end{tabular}} & \textbf{ASRoBERTa} & \textbf{\textit{$\Delta_{domain}$}} & \textbf{Oracle} \\ \hline
\multirow{4}{*}{\textbf{Conversational}} & \textbf{Disfluency}   & 64.23                                                         & 65.18                                                            & 70.52            & 73.13              & 2.61           & 74.18                          \\
                                         & \textbf{Pause}        & 73.67                                                         & 72.45                                                            & 77.31            & 81.29              & 3.98           & 85.20                        \\
                                         & \textbf{Overtalk}     & 80.09                                                         & 81.46                                                            & 83.75            & 85.15              & 1.40           & 95.45                     \\
                                         & \textbf{Question}     & 69.42                                                         & 69.94                                                            & 70.98            & 75.41              & 4.43           & 77.87                    \\ \hline
\multirow{3}{*}{\textbf{Channel}}        & \textbf{Speaker}      & 79.18                                                         & 78.45                                                            & 79.39            & 83.98              & 4.59           & 85.90                    \\
                                         & \textbf{Response-Len} & 63.28                                                         & 64.30                                                            & 65.50            & 68.11              & 2.61           & 70.55                    \\
                                         & \textbf{Turn-Taking}  & 65.26                                                         & 65.04                                                            & 66.23            & 67.45              & 1.22           & 70.36           \\ \hline
\multirow{5}{*}{\textbf{ASR}}            & \textbf{Binary}       & 67.14                                                         & 68.58                                                            & 72.43            & 73.32              & 0.89           & 75.64                       \\
                                         & \textbf{Insertion$^\dagger$}    & 55.66                                                         & 56.49                                                            & 58.99            & 60.87              & 1.88           & 61.73                \\
                                         & \textbf{Deletion$^\dagger$}     & 60.50                                                         & 58.37                                                            & 62.49            & 64.28              & 1.79           & 66.17                  \\
                                         & \textbf{Substitution$^\dagger$} & 48.82                                                         & 49.96                                                            & 50.82            & 55.71              & 4.89           & 54.34                 \\
                                         & \textbf{WER$^\#$}           & 10.38                                                         & 10.09                                                            & 8.50             & 7.42               & -1.08          & 6.27               
\end{tabular}
\caption{Probing results on all tasks. WER$^\#$ is a regression task while others are classification tasks. Tasks with $^\dagger$ are class-wise scores of a multiclass setup. Macro-F1 (higher is better) and MAE (lower is better) are reported for classification \& regression respectively. Negative $\Delta$ for WER is a welcome change since lower WER is better.}
\label{results}
\end{table*}

\subsection{Dataset}
In this work, we focus on evaluating the LM on its ability to encapsulate spoken language understanding in English language with a special focus on real-life spontaneous conversations such as call center interactions. Keeping that in mind, we train ASRoBERTa with transcripts derived from two datasets: a) LibriSpeech \cite{DBLP:conf/icassp/PanayotovCPK15} as a general-purpose ASR dataset (960 hrs.); b) real-life proprietary dataset\footnote{We cannot release the datasets or trained models due to privacy reasons.} (1000 hrs.). All probing setups are carried out on proprietary datasets. For probing tasks, a total
of 10k datapoints are used in the training while validation and
test set comprise of 2k datapoints each. The dataset for each
probe task is class-balanced. It is to be noted that we use real life dataset with audio of 8-16 kHz. The average word-error-rate (WER) of automatic speech recognition (ASR) system for our dataset is 18.38. The probe task dataset creation methodology is explained in Section \ref{creation}.

To evaluate the efficacy of the probing tasks and study the transferability of LM to unseen data, we also evaluate LM fine-tuned with multi-task learning framework on two external datasets: Switchboard Dialog Act (SWDA) \cite{DBLP:journals/lre/CalhounCBMJSB10} and Switchboard Disfluency dataset (SWDB Disfluency) \cite{DBLP:conf/interspeech/ZayatsTWMO19}. SWDA consists of utterances categorized into one of 42 dialog act labels, while SWBD disfluency dataset comes with tokens annotated with disfluency tags. We use the already provided train, test and valid splits of the two datasets to run experiments.

\subsection{Implementation Details}
We use linear probe models \cite{DBLP:conf/iclr/TenneyXCWPMKDBD19, DBLP:conf/iclr/AlainB17} with a cross-entropy loss for classification and mean squared error for regression. The input to probe classifier is the utterance representations derived from \textit{$<$s$>$} for all conversational, channel and WER tasks, while we feed contextualized token embeddings for ASR error type detection in the probing model.
In line with probing task evaluation, we accumulate layer wise results for each task. Each setup is trained for a total of 20 epochs and the best model on the validation set is used to report the results on the test set (Table \ref{results}).

\section{Results and Analysis}

The results for experiments are presented in Table \ref{results} with the improvements obtained from domain pre-training ($\Delta_{domain}$) and then further with task fine-tuning (\textit{Oracle}).

\noindent
\textbf{On Conversation Properties (RQ1a)}: Properties including disfluency, pause, overtalk are captured surprisingly well by domain LM (ASRoBERTa) without explicit supervision for these tasks in the pre-training step. An LM aware of pauses, disfluency and overtalk can understand discourse and speaker segmentation better and hence it is an encouraging sign to note decent results from probing classifiers on the conversational tasks.
The oracle results, although, show that there is a scope for LM to encode these properties even better. A simple downstream task fine-tuning (\textit{Oracle}) provides a considerable boost (2-10\%) in F1-score for these tasks. 

\noindent
\textbf{On Channel Properties (RQ1b)}: The LM when probed on channel properties performs quite well (83.98\%) on speaker role identification. This shows that ASRoBERTa learns to distinguish speaker roles sufficiently in its pre-training step. Although domain pretraining improves the results by 4.5\%, the RoBERTa scores for speaker role identification is close to 80\% indicating that agents and customers play a disntinctive role easily identifiable by the transcript of the speaker itself. On the other side, the results on turn-tasks are below 70\% F1 score. One explanation for this is the inherent difficulty of the turn-tasks which requires a capability to  anticipate a turn-transition and response generation properties such as response-length determined by the contextual understanding of the speaker turns and dialog state. 
Oracle experiment improve over by upto 3\% over ASRoBERTa. Relatively lower oracle scores suggest that a robust mechanism is needed to infuse turn-task properties to make LM aware of contextual information.

\noindent
\textbf{On ASR Properties (RQ1c)}: An LM trained on spoken transcripts is able to distinguish tokens that are erroneous from the correct tokens, scoring 73.32\% F1 on binary classification probe. However, the model is highly inaccurate in distinguishing the error types. The comparisons between ASRoBERTa and RoBERTa on the three error types show that ASRoBERTa is able to identify substitution errors substantially better (55.71\%) than RoBERTa (50.82\%) while has lower gains on insertion and deletion errors ($\Delta_{domain}$ \textit{< 2\%}).
Additionally, ASRoBERTa predicts the WER score of an utterance with a mean absolute error of 7.42 which improves by 1.15 WER once trained with downstream task-finetuning. 
However, Oracle results 
are only marginally better than ASRoBERTa on insertion and deletion. The substitution scores drops down at the cost of improving the error type detection. Thus, LM representations are insufficient to capture ASR error types and pre-training mechanism needs to be revisited to make model learn these ASR properties effectively.

\noindent
\textbf{Meta-Comments on RQ1}: Along with the granular analysis on different properties presented above, the macro-analysis shows some interesting insights:
\begin{itemize}[noitemsep,topsep=0pt]
    \item Default models (\textit{BERT-base, RoBERTa-base}) perform in close-margins on channel properties, as compared to conversational and ASR properties. We hypothesize that conversational and ASR properties require understanding of spontaneous interactions like pauses, stutters and noisy transcriptions which is not present in the base corpus of BERT-base and RoBERTa-base. On the other hand, channel properties are more generic tasks that require lexical understanding (speaker role) and language generation understanding (turn-tasks) which could still be induced from clean texts.
    
    \item A simple pre-training step to induce conversational properties (Chat-RoBERTa) is helpful to realize gains across the tasks. This indicates that we may want to train a better and more suited base-checkpoint for tasks related to dialog systems and spoken language understanding. One of the most simple mechanism would be pre-train a model on publicly available spoken language and/or conversational datasets.
    
    \item We observe significant gains in task-finetuning steps for conversational and channel properties. Hoowever, the absolute results on tasks such as response length and turn-taking may note yet be satisfactory. This calls out a need to identify novel pre-training strategies and/or architecture that could lead to a better base-checkpoint across diverse set of generic tasks related to spoken language understanding.
    
    \item Weak results on ASR probes suggest that MLM objective is insufficient for the pre-trained language models to effectively distinguish between the erroneous tokens and their error types. It would be interesting to try some ASR-specific language model, such as warped language model \cite{DBLP:conf/slt/NamazifarTH21} which is trained with two additional training tasks, namely INSERT and DROP where the model is trained to predict where a random token is inserted into or deleted from the input sequence during training.
    
\end{itemize}

\begin{table}[]
\renewcommand{\arraystretch}{1.15}
\centering
\small
\begin{tabular}{lccc}
\textbf{Relations}       & \textbf{Chat-RoBERTa} & \textbf{ASRoBERTa} & \textbf{$\Delta_{r-asr}$} \\ \hline
all             & 34.9    & 31.4      & -3.5              \\ \hline
\multicolumn{4}{c}{Most frequent relations}               \\ \hline
\textit{prep}   & 66.6    & 62.3      & -4.3              \\
\textit{pobj}   & 71.7    & 60.9      & -10.8             \\
\textit{det}    & 87.1    & 78.8      & -8.3              \\
\textit{nn}     & 74.0      & 71.9      & -2.1              \\
\textit{nsubj}  & 56.4    & 54.1      & -2.3              \\
\textit{amod}   & 82.5    & 79.1      & -3.4              \\
\textit{dobj}   & 78.9    & 75.2      & -3.7              \\
\textit{advmod} & 51.9    & 48.8      & -3.1              \\
\textit{aux}    & 81.4    & 82.1      & 0.7               \\
\textit{num}    & 78.6    & 54.2      & -24.4             \\ \hline
\multicolumn{4}{c}{Relations with highest delta}          \\ \hline
\textit{num}    & 78.6    & 54.2      & -24.4             \\
\textit{ccomp}  & 54.2    & 35.0        & -19.2             \\
\textit{poss}   & 83.9    & 68.4      & -15.5             \\
\textit{conj}   & 59.0      & 43.8      & -15.2             \\
\textit{cc}     & 54.1    & 41.8      & -12.3            
\end{tabular}
\caption{Results (UAS scores) for attention-based probe on dependency parsing. 10 most frequent dependency relations are reported along with the ones with highest delta ($\Delta$).}
\label{dependency}
\end{table}

\noindent
\textbf{On Linguistic Understanding (RQ2)}: The results for attention based probe for dependency parsing is reported in Table \ref{dependency}. For the sake of brevity, we present results for 10 most frequent relations in the dataset. We observe that a language model pre-trained on spoken transcripts (ASRoBERTa) worsens the results obtained by a language model trained on clean text chat (Chat-RoBERTa) by 3.5\% UAS. Additionally, we note that ASRoBERTa performs poor than Chat-RoBERTa in 81.8\% of all dependency relations with \textit{num} relation having the biggest impact on the performance ($\Delta=24.4\%$). ASRoBERTa performs marginally better only on a couple of relations (\textit{aux: $\Delta=0.7\%$, prt: $\Delta=0.5\%$}). These results conclusively show that ASRoBERTa does not retain its understanding of dependency relations leading to huge drop in many relations' performance. A reasoning for this could be the fact that ASRoBERTa is trained on grammatically noisy and non-punctuated texts. With a number of ASR errors and incoherent statements, the model may have learnt to not pay attention to dependency links.

\begin{table}[]
\renewcommand{\arraystretch}{1.15}
\centering
\small
\begin{tabular}{lcc}
\multicolumn{1}{c}{\textbf{Model}}         & \textbf{Entity-Value Acc.} & \textbf{ISA\%} \\ \hline
Chat-RoBERTa                                    & 7.98                     & 3.78            \\
\multicolumn{1}{r}{\textit{+NSP Task}} & 38.62                     & 41.98           \\ \hline
ASRoBERTa                                  & 10.19                     & 3.65            \\
\multicolumn{1}{r}{\textit{+NSP Task}} & 42.64                     & 46.27      
\end{tabular}
\caption{Entity-Value mapping accuracy with inter-sentence attention (ISA\%).}
\label{entity}
\end{table}

In another experiment of entity-value resolution, it is observed that both RoBERTa and ASRoBERTa performs poorly on the task achieving only 7.98\% and 10.19\% respectively on attention-based probing method (Table \ref{entity}). We hypothesize that this could be a result of model's inability to encode the inter-utterance dependencies in a turn-by-turn dyadic conversation. To validate the hypothesis, we perform these steps:
\begin{itemize}[noitemsep,topsep=0pt]
    \item We compute inter-sentence attention score as a percentage (ISA\%) of overall attention that includes self-attention, intra/inter-sentence attention and attention to separator tokens. We observe that the average inter-sentence attention is less than 0.1 except for initial two layers. This shows that model does not put enough emphasis on inter-sentence interactions while encoding the given input. The observation is surprising as model was trained with consecutive dialog turns separated by \textit{</s>} tokens.
    
    \item In order to confirm that poor inter-sentence interaction leads to lower attention and consecutively lower entity-value resolution accuracy, we force the model to learn inter-sentence interactions by training it on a binarized next sentence prediction (NSP) task \cite{DBLP:conf/naacl/DevlinCLT19}. A model would need to understand the input sequence to correctly predict if the last utterance in the input follows the previous utterance. Once the model is trained on NSP task, we run the attention based probing on entity-value resolution once again and note considerable improvements.
\end{itemize}
The experiment with entity-value resolution task demonstrates that LM poorly encodes inter-sentence interactions and thus provides a scope to model these interactions in an effective manner in future works.

\begin{table}[]
\centering
\small
\renewcommand{\arraystretch}{1.15}
\begin{tabular}{lcc}
\textbf{Model}                              & \textbf{SWDA}  & \textbf{\begin{tabular}[c]{@{}l@{}}SWBD Disfluency\end{tabular}} \\ \hline
Chat-RoBERTa                                     & 68.71          & 68.39                                                                     \\
\multicolumn{1}{r}{\textit{\textbf{+ MTL}}} & \textbf{69.72} & \textbf{70.27}                                                            \\ \hline
ASRoBERTa                                   & 67.04          & 72.01                                                                     \\
\multicolumn{1}{r}{\textit{\textbf{+ MTL}}} & \textbf{68.37} & \textbf{75.95} 
\end{tabular}
\caption{Accuracy on SWDA and Switchboard Disfluency dataset in a multi-task learning (MTL) setup}
\vspace{-1.8em}
\label{mtl}
\end{table}


\noindent
\textbf{On properties infused LM: MTL Setup (RQ3)}: 
We show that an LM fine-tuned on all probe tasks in a multi-task learning (MTL) setup performs better than base LM on an unseen corpus (Table \ref{mtl}). MTL improves the accuracy by 1.35\% in SWDA task, while it leads to a gain of 3.94\% on the disfluency task for the ASRoBERTa model. 
Ablation studies across the three categories to train MTL setup show that conversational and channel properties are important for SWDA task while disfluency relies most on conversational properties. 
The results justify our hypothesis that an LM that encodes the conversational, channel and ASR properties is better equipped for downstream tasks in spoken language understanding. The results also demonstrate that the properties learnt by the LM are transferable to unseen SLU tasks.

Another interesting observation is that Chat-RoBERTa outperforms ASRoBERTa on SWDA dataset. Our rationale for this behavior is that SWDA dataset contains majority of utterances in grammatical forms with punctuation, which the real life dataset is devoid of. Higher resemblance of SWDA corpus to chat-dataset could be a reason for difference in performance of the two LMs.

\section{Related Works}
Recent years have seen a surge of works on the theme of neural network interpretability and understandability. A group of work has focused on unveiling secrets of language models \cite{DBLP:clark, DBLP:conf/acl/JawaharSS19}. While \citet{DBLP:clark} probe the surface and linguistic patterns in the attention head of BERT through a set of probing tasks, \citet{DBLP:conf/acl/JawaharSS19} throws light on structural understanding of the language captured by phrase and span representations. Other research works show limitations of LM such as ignoring negation and getting confused by simple distractors \cite{DBLP:conf/acl/KassnerS20} in addition to BERT being inexact in encoding numeracy in its representations \cite{DBLP:conf/emnlp/WallaceWLSG19}. 

There has also been research on probing LM on application specific representations such as question answering \cite{DBLP:conf/cikm/AkenWLG19}, information retrieval \cite{DBLP:conf/emnlp/YilmazYZL19}, recommendation systems \cite{DBLP:conf/recsys/PenhaH20}, dialog systems \cite{DBLP:conf/emnlp/WuX20} etc. The entire spectrum of these works aims to understand the learning capability and properties encoded in the LM along with discovering their shortcomings. The findings in these works have been utilized by the scientific community to create more robust language models: \citet{DBLP:conf/aaai/0001WZLZZZ20} propose Semantics-Aware BERT that performs better than vanilla BERT, while \citet{DBLP:conf/iclr/0225BYWXBPS20} demonstrate that incorporating sentence structure in pre-training LM pushes the results on downstream tasks. 

However, a major section of work have primarily focused
on interpreting LMs trained on clean text. SLU tasks come with their distinctive properties of ASR errors and non-grammatical, ill-punctuated texts. A spoken interaction additionally requires conversational and channel understandability
as highlighted in this work. While works have been carried
out in understanding disfluency \cite{DBLP:conf/aaai/WangCLQLW20, DBLP:conf/interspeech/LinW20} and turn-taking \cite{DBLP:conf/icassp/AldenehDP18, DBLP:conf/interspeech/HaraITK18}, the authors narrowly aim at improving the task specific results by modelling acoustic cues \cite{DBLP:conf/icassp/AldenehDP18, DBLP:conf/icassp/KumarV20} or training with auxiliary tasks \cite{DBLP:conf/icassp/AldenehDP18, DBLP:conf/interspeech/HaraITK18, DBLP:conf/aaai/WangCLQLW20, sundararaman21_interspeech}. The effort in our work is orthogonal to what has been carried out in the past research. We investigate the representations of language model on real-life and benchmark datasets to
identify the strength, limitations and possibility of a generic LM
under the lens of conversational, channel and ASR properties.


\section{Conclusions and Future Work}
We investigate the representations of language model across conversational, channel and ASR properties with probing tasks such as pause, disfluency, overtalk identification, speaker role prediction, turn-tasks and ASR error type. Empirical analysis shows that LM encodes conversational and speaker-type properties to a large extent without external supervision while it has lower performance on turn-tasks and ASR error prediction. Experiments also show that a language model trained on spoken transcripts loses the linguistic understanding of dependency relations. A set of MTL experiments demonstrate the efficacy and transferability of the probe tasks to an unseen SLU dataset that advocates a need and possibility of a domain specific LM.
In future, we would like to research along a few directions that originates from this work: benchmarking properties-infused LM against multimodal counterparts and revisiting the pre-training setup to identify advanced mechanisms to infuse such properties. Additionally, we would also like to evaluate on other language models such as XLNet, GPT etc. to understand the generalizability of the results. 

\bibliographystyle{acl_natbib}
\bibliography{custom}

\begin{thebibliography}{36}
\expandafter\ifx\csname natexlab\endcsname\relax\def\natexlab#1{#1}\fi

\bibitem[{Alain and Bengio(2017)}]{DBLP:conf/iclr/AlainB17}
Guillaume Alain and Yoshua Bengio. 2017.
\newblock \href {https://openreview.net/forum?id=HJ4-rAVtl} {Understanding
  intermediate layers using linear classifier probes}.
\newblock In \emph{5th International Conference on Learning Representations,
  {ICLR} 2017, Toulon, France, April 24-26, 2017, Workshop Track Proceedings}.
  OpenReview.net.

\bibitem[{Aldeneh et~al.(2018)Aldeneh, Dimitriadis, and
  Provost}]{DBLP:conf/icassp/AldenehDP18}
Zakaria Aldeneh, Dimitrios Dimitriadis, and Emily~Mower Provost. 2018.
\newblock Improving end-of-turn detection in spoken dialogues by detecting
  speaker intentions as a secondary task.
\newblock In \emph{2018 {IEEE} International Conference on Acoustics, Speech
  and Signal Processing, {ICASSP} 2018, Calgary, AB, Canada, April 15-20,
  2018}, pages 6159--6163.

\bibitem[{Budzianowski et~al.(2018)Budzianowski, Wen, Tseng, Casanueva, Ultes,
  Ramadan, and Gasic}]{DBLP:conf/emnlp/BudzianowskiWTC18}
Pawel Budzianowski, Tsung{-}Hsien Wen, Bo{-}Hsiang Tseng, I{\~{n}}igo
  Casanueva, Stefan Ultes, Osman Ramadan, and Milica Gasic. 2018.
\newblock \href {https://www.aclweb.org/anthology/D18-1547/} {Multiwoz - {A}
  large-scale multi-domain wizard-of-oz dataset for task-oriented dialogue
  modelling}.
\newblock In \emph{Proceedings of the 2018 Conference on Empirical Methods in
  Natural Language Processing, Brussels, Belgium, October 31 - November 4,
  2018}, pages 5016--5026. Association for Computational Linguistics.

\bibitem[{Calhoun et~al.(2010)Calhoun, Carletta, Brenier, Mayo, Jurafsky,
  Steedman, and Beaver}]{DBLP:journals/lre/CalhounCBMJSB10}
Sasha Calhoun, Jean Carletta, Jason~M. Brenier, Neil Mayo, Dan Jurafsky, Mark
  Steedman, and David Beaver. 2010.
\newblock The nxt-format switchboard corpus: a rich resource for investigating
  the syntax, semantics, pragmatics and prosody of dialogue.
\newblock \emph{Lang. Resour. Evaluation}, 44(4):387--419.

\bibitem[{Clark et~al.(2019)Clark, Khandelwal, Levy, and Manning}]{DBLP:clark}
Kevin Clark, Urvashi Khandelwal, Omer Levy, and Christopher~D. Manning. 2019.
\newblock \href {http://arxiv.org/abs/1906.04341} {What does {BERT} look at? an
  analysis of bert's attention}.
\newblock \emph{ACL Workshop BlackboxNLP: Analyzing and Interpreting Neural
  Networks for NLP}.

\bibitem[{Conneau et~al.(2018)Conneau, Kruszewski, Lample, Barrault, and
  Baroni}]{DBLP:conf/acl/BaroniBLKC18}
Alexis Conneau, Germ{\'{a}}n Kruszewski, Guillaume Lample, Lo{\"{\i}}c
  Barrault, and Marco Baroni. 2018.
\newblock \href {https://doi.org/10.18653/v1/P18-1198} {What you can cram into
  a single {\textbackslash}{\textdollar}{\&}!{\#}* vector: Probing sentence
  embeddings for linguistic properties}.
\newblock In \emph{Proceedings of the 56th Annual Meeting of the Association
  for Computational Linguistics, {ACL} 2018, Melbourne, Australia, July 15-20,
  2018, Volume 1: Long Papers}, pages 2126--2136. Association for Computational
  Linguistics.

\bibitem[{Devlin et~al.(2019)Devlin, Chang, Lee, and
  Toutanova}]{DBLP:conf/naacl/DevlinCLT19}
Jacob Devlin, Ming{-}Wei Chang, Kenton Lee, and Kristina Toutanova. 2019.
\newblock {BERT:} pre-training of deep bidirectional transformers for language
  understanding.
\newblock In \emph{Proceedings of the 2019 Conference of the North American
  Chapter of the Association for Computational Linguistics: Human Language
  Technologies, {NAACL-HLT} 2019, MN, USA, June 2-7, 2019}, pages 4171--4186.

\bibitem[{Dinkar et~al.(2020)Dinkar, Vasilescu, Pelachaud, and
  Clavel}]{DBLP:conf/icassp/DinkarVPC20}
Tanvi Dinkar, Ioana Vasilescu, Catherine Pelachaud, and Chlo{\'{e}} Clavel.
  2020.
\newblock How confident are you? exploring the role of fillers in the automatic
  prediction of a speaker's confidence.
\newblock In \emph{2020 {IEEE} International Conference on Acoustics, Speech
  and Signal Processing, {ICASSP} 2020, Barcelona, Spain, May 4-8, 2020}, pages
  8104--8108.

\bibitem[{Georgiou et~al.(2019)Georgiou, Papaioannou, and
  Potamianos}]{DBLP:conf/interspeech/GeorgiouPP19}
Efthymios Georgiou, Charilaos Papaioannou, and Alexandros Potamianos. 2019.
\newblock Deep hierarchical fusion with application in sentiment analysis.
\newblock In \emph{Interspeech 2019, 20th Annual Conference of the
  International Speech Communication Association, Graz, Austria, 15-19
  September 2019}, pages 1646--1650.

\bibitem[{Hara et~al.(2018)Hara, Inoue, Takanashi, and
  Kawahara}]{DBLP:conf/interspeech/HaraITK18}
Kohei Hara, Koji Inoue, Katsuya Takanashi, and Tatsuya Kawahara. 2018.
\newblock Prediction of turn-taking using multitask learning with prediction of
  backchannels and fillers.
\newblock In \emph{Interspeech 2018, 19th Annual Conference of the
  International Speech Communication Association, Hyderabad, India, 2-6
  September 2018}, pages 991--995.

\bibitem[{Jawahar et~al.(2019)Jawahar, Sagot, and
  Seddah}]{DBLP:conf/acl/JawaharSS19}
Ganesh Jawahar, Beno{\^{\i}}t Sagot, and Djam{\'{e}} Seddah. 2019.
\newblock What does {BERT} learn about the structure of language?
\newblock In \emph{Proceedings of the 57th Conference of the Association for
  Computational Linguistics, {ACL} 2019, Florence, Italy}, pages 3651--3657.

\bibitem[{Kassner and Sch{\"{u}}tze(2020)}]{DBLP:conf/acl/KassnerS20}
Nora Kassner and Hinrich Sch{\"{u}}tze. 2020.
\newblock Negated and misprimed probes for pretrained language models: Birds
  can talk, but cannot fly.
\newblock In \emph{Proceedings of the 58th Annual Meeting of the Association
  for Computational Linguistics, {ACL}}, pages 7811--7818.

\bibitem[{Kim et~al.(2020)Kim, Jeong, and Lee}]{kim2020speaker}
Jonggu Kim, Yewon Jeong, and Jong-Hyeok Lee. 2020.
\newblock Speaker-informed time-and-content-aware attention for spoken language
  understanding.
\newblock \emph{Computer Speech \& Language}, 60:101022.

\bibitem[{Kumar and Vepa(2020)}]{DBLP:conf/icassp/KumarV20}
Ayush Kumar and Jithendra Vepa. 2020.
\newblock Gated mechanism for attention based multi modal sentiment analysis.
\newblock In \emph{2020 {IEEE} International Conference on Acoustics, Speech
  and Signal Processing, {ICASSP} 2020, Barcelona, Spain}, pages 4477--4481.

\bibitem[{Li et~al.(2020)Li, Zhou, Li, Xu, and
  Cao}]{DBLP:journals/corr/abs-2012-15150}
Zhongli Li, Qingyu Zhou, Chao Li, Ke~Xu, and Yunbo Cao. 2020.
\newblock \href {http://arxiv.org/abs/2012.15150} {Improving {BERT} with
  syntax-aware local attention}.
\newblock \emph{CoRR}, abs/2012.15150.

\bibitem[{Lin and Wang(2020)}]{DBLP:conf/interspeech/LinW20}
Binghuai Lin and Liyuan Wang. 2020.
\newblock Joint prediction of punctuation and disfluency in speech transcripts.
\newblock In \emph{Interspeech 2020, 21st Annual Conference of the
  International Speech Communication Association, Virtual Event, Shanghai,
  China}, pages 716--720.

\bibitem[{Liu et~al.(2019)Liu, Ott, Goyal, Du, Joshi, Chen, Levy, Lewis,
  Zettlemoyer, and Stoyanov}]{DBLP:journals/corr/abs-1907-11692}
Yinhan Liu, Myle Ott, Naman Goyal, Jingfei Du, Mandar Joshi, Danqi Chen, Omer
  Levy, Mike Lewis, Luke Zettlemoyer, and Veselin Stoyanov. 2019.
\newblock \href {http://arxiv.org/abs/1907.11692} {Roberta: {A} robustly
  optimized {BERT} pretraining approach}.
\newblock \emph{CoRR}, abs/1907.11692.

\bibitem[{Marcus et~al.(1993)Marcus, Santorini, and
  Marcinkiewicz}]{DBLP:journals/coling/MarcusSM94}
Mitchell~P. Marcus, Beatrice Santorini, and Mary~Ann Marcinkiewicz. 1993.
\newblock Building a large annotated corpus of english: The penn treebank.
\newblock \emph{Comput. Linguistics}, 19(2):313--330.

\bibitem[{Namazifar et~al.(2021)Namazifar, T{\"{u}}r, and
  Hakkani{-}T{\"{u}}r}]{DBLP:conf/slt/NamazifarTH21}
Mahdi Namazifar, G{\"{o}}khan T{\"{u}}r, and Dilek Hakkani{-}T{\"{u}}r. 2021.
\newblock \href {https://doi.org/10.1109/SLT48900.2021.9383493} {Warped
  language models for noise robust language understanding}.
\newblock In \emph{{IEEE} Spoken Language Technology Workshop, {SLT} 2021,
  Shenzhen, China, January 19-22, 2021}, pages 981--988. {IEEE}.

\bibitem[{Panayotov et~al.(2015)Panayotov, Chen, Povey, and
  Khudanpur}]{DBLP:conf/icassp/PanayotovCPK15}
Vassil Panayotov, Guoguo Chen, Daniel Povey, and Sanjeev Khudanpur. 2015.
\newblock \href {https://doi.org/10.1109/ICASSP.2015.7178964} {Librispeech: An
  {ASR} corpus based on public domain audio books}.
\newblock In \emph{2015 {IEEE} International Conference on Acoustics, Speech
  and Signal Processing, {ICASSP} 2015, South Brisbane, Queensland, Australia,
  April 19-24, 2015}, pages 5206--5210. {IEEE}.

\bibitem[{Penha and Hauff(2020)}]{DBLP:conf/recsys/PenhaH20}
Gustavo Penha and Claudia Hauff. 2020.
\newblock What does {BERT} know about books, movies and music? probing {BERT}
  for conversational recommendation.
\newblock In \emph{RecSys 2020: Fourteenth {ACM} Conference on Recommender
  Systems, Virtual Event, Brazil, September 22-26, 2020}, pages 388--397.

\bibitem[{Roddy et~al.(2018)Roddy, Skantze, and
  Harte}]{DBLP:conf/interspeech/RoddySH18}
Matthew Roddy, Gabriel Skantze, and Naomi Harte. 2018.
\newblock Investigating speech features for continuous turn-taking prediction
  using lstms.
\newblock In \emph{Interspeech 2018, 19th Annual Conference of the
  International Speech Communication Association, Hyderabad, India, 2-6
  September 2018}, pages 586--590.

\bibitem[{Sundararaman et~al.(2019)Sundararaman, Subramanian, Wang, Si, Shen,
  Wang, and Carin}]{DBLP:journals/corr/abs-1911-06156}
Dhanasekar Sundararaman, Vivek Subramanian, Guoyin Wang, Shijing Si, Dinghan
  Shen, Dong Wang, and Lawrence Carin. 2019.
\newblock \href {http://arxiv.org/abs/1911.06156} {Syntax-infused transformer
  and {BERT} models for machine translation and natural language
  understanding}.
\newblock \emph{CoRR}, abs/1911.06156.

\bibitem[{Sundararaman et~al.(2021)Sundararaman, Kumar, and
  Vepa}]{sundararaman21_interspeech}
Mukuntha~Narayanan Sundararaman, Ayush Kumar, and Jithendra Vepa. 2021.
\newblock \href {https://doi.org/10.21437/Interspeech.2021-1582} {{PhonemeBERT:
  Joint Language Modelling of Phoneme Sequence and ASR Transcript}}.
\newblock In \emph{Proc. Interspeech 2021}, pages 3236--3240.

\bibitem[{Tenney et~al.(2019)Tenney, Xia, Chen, Wang, Poliak, McCoy, Kim,
  Durme, Bowman, Das, and Pavlick}]{DBLP:conf/iclr/TenneyXCWPMKDBD19}
Ian Tenney, Patrick Xia, Berlin Chen, Alex Wang, Adam Poliak, R.~Thomas McCoy,
  Najoung Kim, Benjamin~Van Durme, Samuel~R. Bowman, Dipanjan Das, and Ellie
  Pavlick. 2019.
\newblock What do you learn from context? probing for sentence structure in
  contextualized word representations.
\newblock In \emph{7th International Conference on Learning Representations,
  {ICLR} 2019, New Orleans, LA, USA}.

\bibitem[{van Aken et~al.(2019)van Aken, Winter, L{\"{o}}ser, and
  Gers}]{DBLP:conf/cikm/AkenWLG19}
Betty van Aken, Benjamin Winter, Alexander L{\"{o}}ser, and Felix~A. Gers.
  2019.
\newblock How does {BERT} answer questions?: {A} layer-wise analysis of
  transformer representations.
\newblock In \emph{Proceedings of the 28th {ACM} International Conference on
  Information and Knowledge Management, {CIKM} 2019, Beijing, China}, pages
  1823--1832.

\bibitem[{Wallace et~al.(2019)Wallace, Wang, Li, Singh, and
  Gardner}]{DBLP:conf/emnlp/WallaceWLSG19}
Eric Wallace, Yizhong Wang, Sujian Li, Sameer Singh, and Matt Gardner. 2019.
\newblock Do {NLP} models know numbers? probing numeracy in embeddings.
\newblock In \emph{Proceedings of the 2019 Conference on Empirical Methods in
  Natural Language Processing and the 9th International Joint Conference on
  Natural Language Processing, {EMNLP-IJCNLP} 2019, Hong Kong, China}, pages
  5306--5314.

\bibitem[{Wang et~al.(2020{\natexlab{a}})Wang, Che, Liu, Qin, Liu, and
  Wang}]{DBLP:conf/aaai/WangCLQLW20}
Shaolei Wang, Wanxiang Che, Qi~Liu, Pengda Qin, Ting Liu, and William~Yang
  Wang. 2020{\natexlab{a}}.
\newblock Multi-task self-supervised learning for disfluency detection.
\newblock In \emph{The Thirty-Fourth {AAAI} Conference on Artificial
  Intelligence, {AAAI} 2020, The Thirty-Second Innovative Applications of
  Artificial Intelligence Conference, {IAAI} 2020, The Tenth {AAAI} Symposium
  on Educational Advances in Artificial Intelligence, {EAAI} 2020, New York,
  NY, USA, February 7-12, 2020}, pages 9193--9200.

\bibitem[{Wang et~al.(2020{\natexlab{b}})Wang, Bi, Yan, Wu, Xia, Bao, Peng, and
  Si}]{DBLP:conf/iclr/0225BYWXBPS20}
Wei Wang, Bin Bi, Ming Yan, Chen Wu, Jiangnan Xia, Zuyi Bao, Liwei Peng, and
  Luo Si. 2020{\natexlab{b}}.
\newblock Structbert: Incorporating language structures into pre-training for
  deep language understanding.
\newblock In \emph{8th International Conference on Learning Representations,
  {ICLR} 2020, Addis Ababa, Ethiopia}.

\bibitem[{Weiner et~al.(2017)Weiner, Engelbart, and
  Schultz}]{DBLP:conf/interspeech/WeinerES17}
Jochen Weiner, Mathis Engelbart, and Tanja Schultz. 2017.
\newblock Manual and automatic transcriptions in dementia detection from
  speech.
\newblock In \emph{Interspeech 2017, 18th Annual Conference of the
  International Speech Communication Association, Stockholm, Sweden, August
  20-24, 2017}, pages 3117--3121.

\bibitem[{Wu and Xiong(2020)}]{DBLP:conf/emnlp/WuX20}
Chien{-}Sheng Wu and Caiming Xiong. 2020.
\newblock Probing task-oriented dialogue representation from language models.
\newblock In \emph{Proceedings of the 2020 Conference on Empirical Methods in
  Natural Language Processing, {EMNLP} 2020, Online, November 16-20, 2020},
  pages 5036--5051.

\bibitem[{Yilmaz et~al.(2019)Yilmaz, Yang, Zhang, and
  Lin}]{DBLP:conf/emnlp/YilmazYZL19}
Zeynep~Akkalyoncu Yilmaz, Wei Yang, Haotian Zhang, and Jimmy Lin. 2019.
\newblock Cross-domain modeling of sentence-level evidence for document
  retrieval.
\newblock In \emph{Proceedings of the 2019 Conference on Empirical Methods in
  Natural Language Processing and the 9th International Joint Conference on
  Natural Language Processing, {EMNLP-IJCNLP} 2019, Hong Kong, China, November
  3-7, 2019}, pages 3488--3494.

\bibitem[{Zayats et~al.(2019)Zayats, Tran, Wright, Mansfield, and
  Ostendorf}]{DBLP:conf/interspeech/ZayatsTWMO19}
Vicky Zayats, Trang Tran, Richard~A. Wright, Courtney Mansfield, and Mari
  Ostendorf. 2019.
\newblock Disfluencies and human speech transcription errors.
\newblock In \emph{Interspeech 2019, 20th Annual Conference of the
  International Speech Communication Association, Graz, Austria, 15-19
  September 2019}, pages 3088--3092.

\bibitem[{Zhang et~al.(2018{\natexlab{a}})Zhang, Lee, Polymenakos, and
  Radev}]{DBLP:conf/aaai/ZhangLPR18}
Rui Zhang, Honglak Lee, Lazaros Polymenakos, and Dragomir~R. Radev.
  2018{\natexlab{a}}.
\newblock Addressee and response selection in multi-party conversations with
  speaker interaction rnns.
\newblock In \emph{Proceedings of the Thirty-Second {AAAI} Conference on
  Artificial Intelligence, (AAAI-18), the 30th innovative Applications of
  Artificial Intelligence (IAAI-18), and the 8th {AAAI} Symposium on
  Educational Advances in Artificial Intelligence (EAAI-18), New Orleans,
  Louisiana, USA, February 2-7, 2018}, pages 5690--5697.

\bibitem[{Zhang et~al.(2018{\natexlab{b}})Zhang, Dinan, Urbanek, Szlam, Kiela,
  and Weston}]{DBLP:conf/acl/KielaWZDUS18}
Saizheng Zhang, Emily Dinan, Jack Urbanek, Arthur Szlam, Douwe Kiela, and Jason
  Weston. 2018{\natexlab{b}}.
\newblock \href {https://doi.org/10.18653/v1/P18-1205} {Personalizing dialogue
  agents: {I} have a dog, do you have pets too?}
\newblock In \emph{Proceedings of the 56th Annual Meeting of the Association
  for Computational Linguistics, {ACL} 2018, Melbourne, Australia, July 15-20,
  2018, Volume 1: Long Papers}, pages 2204--2213. Association for Computational
  Linguistics.

\bibitem[{Zhang et~al.(2020)Zhang, Wu, Zhao, Li, Zhang, Zhou, and
  Zhou}]{DBLP:conf/aaai/0001WZLZZZ20}
Zhuosheng Zhang, Yuwei Wu, Hai Zhao, Zuchao Li, Shuailiang Zhang, Xi~Zhou, and
  Xiang Zhou. 2020.
\newblock Semantics-aware {BERT} for language understanding.
\newblock In \emph{The Thirty-Fourth {AAAI} Conference on Artificial
  Intelligence, {AAAI} 2020, The Thirty-Second Innovative Applications of
  Artificial Intelligence Conference, {IAAI} 2020, 10th {AAAI} Symposium on
  Educational Advances in Artificial Intelligence, {EAAI} 2020, New York, NY,
  USA, February 7-12, 2020}, pages 9628--9635.

\end{thebibliography}

\appendix

\section{Appendix}
\label{sec:appendix}

\subsection{Dataset Creation}
\label{creation}

In this section, we describe the methodology used to curate the dataset for each probing tasks mentioned in Section \ref{probing}.

\subsubsection{Conversational Properties}

\begin{itemize}
    \item \textbf{Disfluency Detection}: To create the dataset for disfluency we follow two-step process: i) candidate retrieval, ii) manual annotation. For candidate retrieval, we have a couple of pipelines, one relying on lexical cues and another taking hints from the speech cues. For lexical cues, we use a simple keyphrase based lookup to identify the candidate speaker utterances containing markers for disfluency such as repetition of words, filler words and discourse markers. We, however, note that lexical cues only forms a subset of all disfluencies. To include speech induced disfluencies, we identify turns that contains multiple intermittent pauses within a short time frame. This candidate extraction ensures that we are probing a lexical model on its capability to understand disfluencies that are not keyword based and hence a greater semantic understanding is needed to locate such disfluent contexts in the spoken transcripts. Once such candidates are obtained, we use manual annotators to label them into fluent and disfluent categories, thus generating a binary classification dataset.
    
    \item \textbf{Pause Identification}: The dataset for pause identification is generated by retrieving all utterances where speaker takes one or more pauses longer than 5 seconds in an individual turn. Additionally, we exclude all such utterances that could be a disfluent candidate. The rationale behind such decision is to ensure that we not add bias in the pause identification probe setup such that disfluent markers lead to a better pause understanding. In this case, there is no need for manual annotation as the pause duration obtained from the speech information acts as gold label for the utterances. We frame pause identification as a binary classification where one class denotes that speaker takes a pause longer than 5 sec. in a given turn of the conversation while other class denotes the absence of pauses in the turn.
    
    \item \textbf{Overtalk Detection}: An overtalk in a given dual speaker conversation would like Table \ref{overtalk}. These leads to broken and incomprehensible turns compared to their clean versions of transcripts. While it is easy to detect such overtalk when the recording lines itself are unique to speakers, this is an arduous task for monochannel recordings. Thus, we curiously probe the language model to see if it is able to detect transcripts with overtalks. We follow a simple mechanism to curate overtalk dataset: we identify overtalk via rule based system from dual channel call and create corresponding monochannel transcripts (AB-mono in Table \ref{overtalk}). These monochannel transcripts are dataset that belong to `overtalk' label. Any non-overtalk transcript from a single speaker belong to `non-overtalk' label. Thus, overtalk detection too is a binary classification task.
    
    \begin{table}[h]
    \renewcommand{\arraystretch}{1.15}
    \small
    \centering
\begin{tabular}{ll}
\textbf{Speaker} & \textbf{Transcript}              \\
A                & i'm                              \\
B                & yes                              \\
A                & not                              \\
B                & i know                           \\
A                & referring to                     \\
B                & how                              \\
A                & transaction                      \\
B                & it works                         \\ \hline
A - clean        & \textit{i'm not referring to transaction} \\
B - clean        & \textit{yes i know how it works}         \\ \hline
AB-mono         & \textit{i'm yes not i know referring to how} \\
                & \textit{transaction it works} \\

\end{tabular}
\caption{An example of a dual channel transcript containing overtalk. A-clean, B-clean refers to rewriting what A and B's transcript would look like if had they not spoken over each other.}
\label{overtalk}
\end{table}

    \item \textbf{Question Identification}: Question identification is a non-trivial problem in spoken language transcripts due to inaccurate predictions and non-surface question constructs, such as: \textit{and you would like me to continue}. In the example, the speaker intends to actually ask a question if he/she can continue further. To create question dataset, we follow a similar two step process: i) candidate generation; and ii) manual annotation. For candidate generation, we two approaches: a) question keyphrases based lookup; and b) using speaker replies that indicate presence of boolen answers such as \textit{yes that's true} or entities such as \textit{sure, it would be ninety five dollars}. Once candidate turns are collected, we ask manual annotators to mark the question span and the category of the question.
        
        
        
    
\end{itemize}

\subsubsection{Channel Properties}

\begin{itemize}
    \item \textbf{Speaker Role}: We restrict the domain of the dataset to call center conversations with two speaker roles: a) agent and b) customer. These two speakers have distinctive properties as per their role, agent are expected to be fluent while being empathetic and enthusiastic towards the customers. Customers, on the other hand, are calling to raise their concerns or ask for help with many cases of heated engagements with agents. Thus, we create a speaker role binary classification dataset with two classes: agent and customer obtained directly from dual channel calls.
    
    \item \textbf{Response-Length}: To create dataset for this task, we use the time-duration of an utterance's response to categorize it into a short response if it is is less than 30 seconds of duration. For response lasting longer than 30 seconds are considered to be long response. The utterance is what is taken as consideration in the dataset with the label being short or long depending on the utterance's response length. To arrive at the duration, we discard turns in the initial 5\%ile and last 5\%ile category based on the turn duration in order to remove outliers. For the remaining dataset, we choose the time period that belongs to the 50th percentile.
    
    \item \textbf{Turn-Taking}: We pose the task as a binary classification task where a label of `turn-continue' would mean that with the given utterance, the speaker has not yet finished speaking and hence would add on further, while a label of `turn-break' would refer to the scenario where speaker has finished its turn of speaking.
\end{itemize}

\subsubsection{ASR Properties}

\begin{itemize}
    \item \textbf{Error Type Prediction}: We obtain spoken transcripts from automatic speech recognition system. Additionally, we get manual transcripts for the same set of calls. We run alignment over the parallel corpus to identify the categories of errors for each token in spoken transcripts. We formulate two tasks for error type prediction: binary classification where the model only needs to predict if the token under consideration is erroneous; and multiclass classification in which for any erroneous token, the model needs to predict the error type: insertion, deletion or substitution.
    
    \item \textbf{WER Prediction}: From the dataset obtained in the above task, we compute the word error rate for the spoken transcript as compared to manual transcripts. A regression model is trained for this task.

\end{itemize}

\subsection{Token Baseline}

In addition to the three setups present in Section \ref{exp}, we also evaluate a ngram based token baseline to contrast the performance observed from language model based probe classifiers. We use ngram features upto quadgrams as input to the model. We use logistic and linear regression for all classification and regression tasks respectively. The results for token baseline is presented in Table \ref{token}.

\begin{table}[h]
\centering
\begin{tabular}{l|c}
\textbf{Probe Tasks} & Token Baseline \\ \hline
Pause                & 70.05          \\
Disfluency           & 66.55          \\
Overtalk             & 60.94          \\
Question             & 69.49          \\ \hline
Speaker              & 79.75          \\
Response-Length      & 65.25          \\
Turn-Taking          & 58.29          \\ \hline
Binary               & 54.34          \\
Insertion            & 50.29          \\
Deletion             & 49.26          \\
Substitution         & 41.33          \\
WER                  & 10.31         
\end{tabular}
\caption{}
\label{token}
\end{table}

\subsection{Attention Scores}

\begin{figure*}
\centering
\begin{subfigure}[b]{0.7\textwidth}
   \includegraphics[width=1\linewidth]{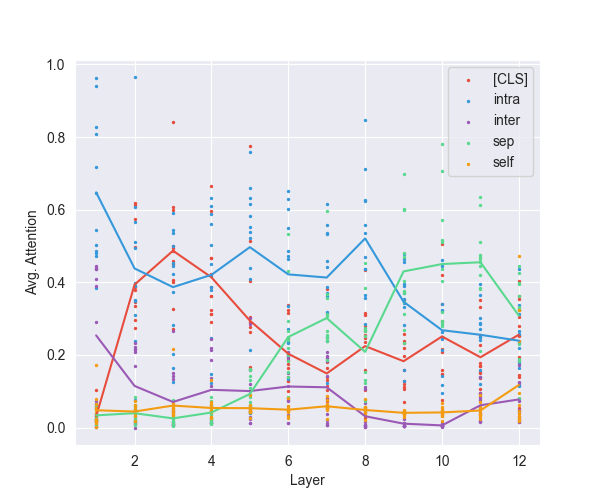}
   \caption{Average attention scores for ASRoBERTa across all layers. The graph shows that inter-sentence attention (ISA) is low for all layers except the initial layer. \textit{[CLS]} and \textit{sep} corresponds to \textit{<s>} and \textit{</s>} in ASRoBERTa.}
   \label{asr} 
\end{subfigure}

\begin{subfigure}[b]{0.7\textwidth}
   \includegraphics[width=1\linewidth]{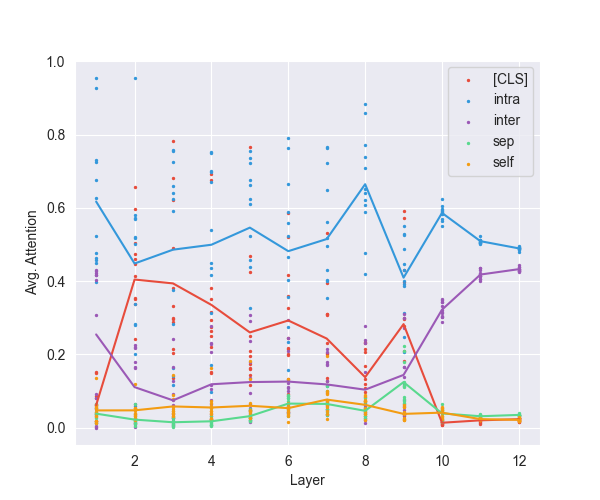}
   \caption{Average attention scores across all layers for ASRoBERTa trained on next-sentence-prediction (NSP) task. The graph shows that inter-sentence attention (ISA) goes up, especially for final layer suggesting that inter-sentence understanding is needed for solving NSP task. \textit{[CLS]} and \textit{sep} corresponds to \textit{<s>} and \textit{</s>} in ASRoBERTa.}
   \label{asrnsp}
\end{subfigure}
\label{attn_graph}
\end{figure*}

In this section, we show the variation in average attention scores between a pair of tokens across all layers for ASRoBERTa model trained only on masked language modelling (MLM) task and ASRoBERTa-NSP model which is further fine-tuned on next-sentence-prediction (NSP) task. The graphs demonstrate that inter-sentence scores rise in final layers (Figure \ref{asrnsp}) indicating the need of inter-sentence interaction for solving NSP task. The code for graph generation has been adapted from the paper by \citet{DBLP:clark}.

\end{document}